\documentclass[10pt,twocolumn,letterpaper]{article}

\usepackage{cuted}    %
\usepackage{caption}  %
\usepackage{booktabs}   %
\usepackage{multirow}   %
\usepackage{graphicx}   %
\usepackage{caption}    %
\usepackage[dvipsnames]{xcolor}     %
\usepackage{colortbl}   %
\usepackage[pagenumbers]{cvpr} %

\definecolor{bestblue}{RGB}{130, 180, 240}     %
\definecolor{secondblue}{RGB}{182, 212, 248}   %
\definecolor{thirdblue}{RGB}{222, 237, 253}    %
\newcommand{\besthl}[1]{{\setlength{\fboxsep}{1.5pt}\colorbox{bestblue}{\textbf{#1}}}}
\newcommand{\secondbesthl}[1]{{\setlength{\fboxsep}{1.5pt}\colorbox{secondblue}{#1}}}

\definecolor{oursrow}{RGB}{238, 240, 243}
\newcommand{\oursrow}{\rowcolor{oursrow}}

\definecolor{refgray}{RGB}{130, 130, 130}
\newcommand{\rg}[1]{\textcolor{refgray}{#1}}

\definecolor{cvprblue}{rgb}{0.21,0.49,0.74}
\usepackage[pagebackref,breaklinks,colorlinks,allcolors=cvprblue]{hyperref}

\AtEndPreamble{
    \crefname{table}{Table}{Tables}
    \crefname{subsection}{Sec.}{Secs.}
    \Crefname{subsection}{Section}{Sections}
}

\title{StreamTAP3D: Streaming 3D Point Tracking with Hybrid Memory}

\author{
Zhuoyuan Li$^{1*}$ \quad
Jiahao Lu$^{2*}$ \quad
Andrea Emir Sevincel$^{2}$ \quad
Yanzhe Liang$^{1}$ \quad
Chengfeng Zhao$^{2}$ \\
Haiping Wang$^{2}$ \quad
Yu Lu$^{3\dagger}$ \quad
Zijiao Zeng$^{3\dagger}$ \quad
Yuan Liu$^{2\ddagger}$ \quad
Bing Wang$^{1\ddagger}$ \\[0.4em]
{\small $^{1}$The Hong Kong Polytechnic University \quad
$^{2}$The Hong Kong University of Science and Technology \quad
$^{3}$Tencent Games Motus} \\[0.4em]
{\normalsize \url{https://charlie839242.github.io/StreamTAP3D-page/}}
}

\begin{document}
\maketitle
{\let\thefootnote\relax\footnotetext{$^*$Equal contribution.}\footnotetext{$^\dagger$Project lead.}\footnotetext{$^\ddagger$Corresponding authors.}}

\begin{strip}
  \centering
  \vspace{-4.2em}
  \includegraphics[width=\linewidth]{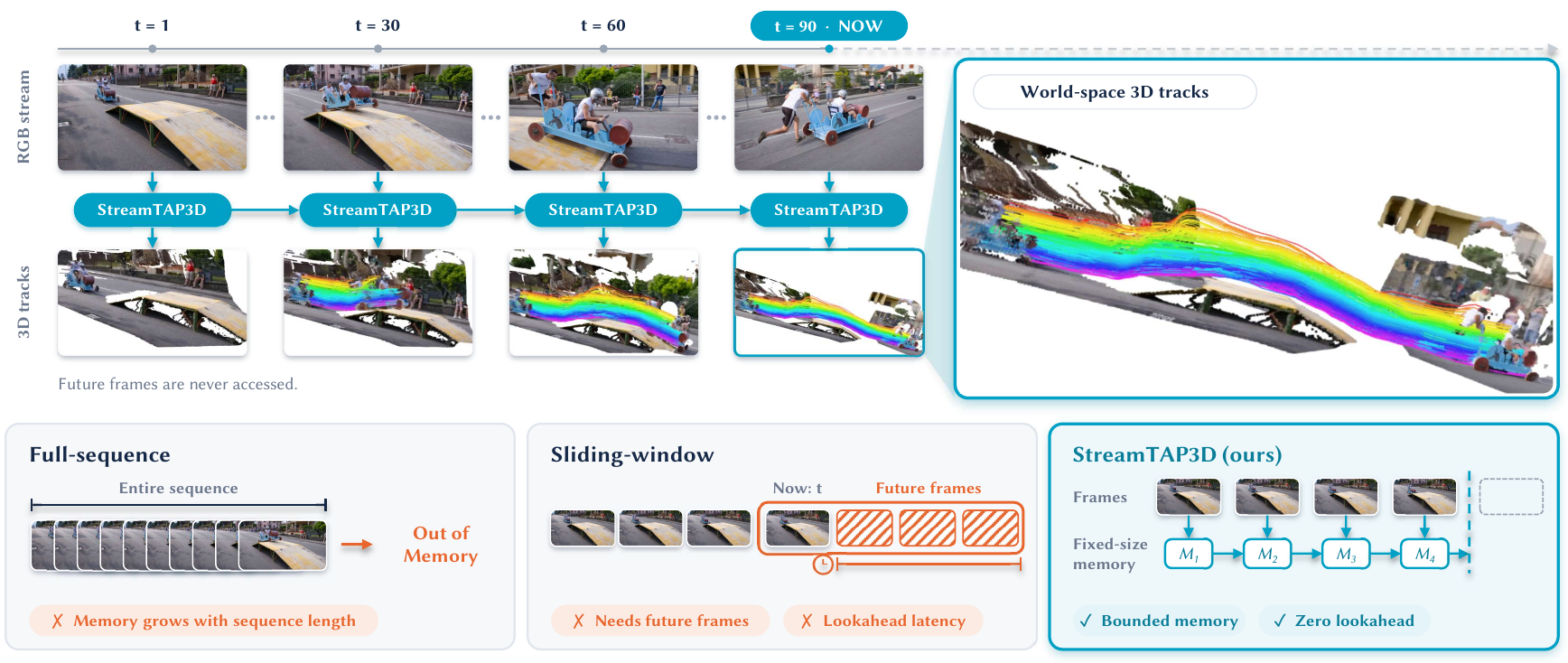}
  \captionof{figure}{\textbf{Streaming 3D Point Tracking.} 
\textbf{Top:} Given a stream of RGB images as input, our system performs simultaneous 3D reconstruction and 3D point tracking in a streaming manner, estimating per-frame camera parameters, 3D geometry, and 3D point trajectories with each incoming frame. 
\textbf{Bottom:} Unlike \emph{full-sequence} trackers or \emph{sliding-window} trackers, our approach processes each frame upon arrival, achieving zero lookahead. Operating at 17 FPS within a 10 GB memory budget ($384 \times 512$ resolution, 256 tracked points), our method handles complex camera motion and dynamic scenes with constant per-frame cost regardless of video length.}
  \label{fig:teaser}
\end{strip}

\begin{abstract}
Perceiving 3D motion in live video calls for point tracking that responds to each frame as it arrives, without waiting for future observations. Many existing methods rely on offline full-video processing or multi-frame sliding windows, incurring lookahead latency or memory costs that grow with video length. We present StreamTAP3D, a streaming framework that tracks 3D query points from monocular RGB video strictly causally under a bounded memory budget. To capture temporal context, we introduce a hybrid long-short-term streaming memory that compresses long-term trajectory history into constant-size states while retaining fine-grained short-term motion context. To stabilize 3D refinement under noisy online geometry, we design a coupled 2D-3D streaming architecture that uses geometry-independent 2D predictions to guide 3D iterative refinement and promote 2D-3D consistency, while letting the 3D branch benefit from abundant 2D tracking data. Given $384 \times 512$ video streams, StreamTAP3D tracks 256 points at 17 FPS under 10 GB of VRAM, including the cost of streaming reconstruction. Extensive experiments across four 3D tracking datasets demonstrate that StreamTAP3D, with zero lookahead latency, remains competitive with offline and sliding-window methods that access future frames.
\end{abstract}
    
\section{Introduction}
\label{sec:intro}

Tracking how points move in 3D over time underpins a wide range of applications, from robotic manipulation to augmented reality~\cite{Koppula2024TAPVid3DAB, Xiao2024SpatialTrackerTA, Lu2026Track4WorldFW, Karaev2025CoTracker3SA, Ngo2024DELTADE, Liu2025TraceAR, Zhang2025EfficientlyRD, Xiao2025SpatialTrackerV23P}. In many of these settings, tracking must keep pace with the incoming video: a robot reacting to a moving object or an AR headset anchoring content to a moving surface cannot afford to wait for future frames. Achieving such responsiveness requires a 3D point tracker that processes each frame as it arrives, with bounded memory.

Despite rapid progress in 2D point tracking~\cite{Karaev2023CoTrackerII, Karaev2025CoTracker3SA, Harley2022ParticleVR, Doersch2023TAPIRTA, Harley2025AllTrackerED, Lai2026CoWTrackerTB} and its extension to 3D~\cite{Xiao2024SpatialTrackerTA,Ngo2024DELTADE, Lu2026Track4WorldFW, Zhang2025EfficientlyRD, Zhang2025TAPIP3DTA}, existing 3D trackers fall short of this requirement. Offline methods~\cite{Zhang2025EfficientlyRD, Xiao2025SpatialTrackerV23P, Luo20264RC4R, Jiang2026OmniXAA, Sucar2026VDPM4V, Cheng2026MoSE3LW, Karhade2025Any4DUF} process the entire video, which precludes live streaming and scales poorly with video length. To handle longer videos, sliding-window methods~\cite{Zhang2025TAPIP3DTA, Lu2026Track4WorldFW, Ngo2024DELTADE, Xiao2024SpatialTrackerTA, Wang2024SceneTrackerLS} propagate tracks across overlapping multi-frame windows (e.g., $8$ or $16$ frames). Though often termed ``online'', they must buffer future frames before producing outputs for the current frame, incurring lookahead latency (\cref{fig:teaser}, bottom). Moreover, many assume precomputed depth or camera poses~\cite{Zhang2025TAPIP3DTA, Ngo2024DELTADE, Xiao2024SpatialTrackerTA, Wang2024SceneTrackerLS}, overlooking the cost of estimating geometry on the fly.

In this paper, we ask: \emph{Can we perform accurate 3D point tracking in a streaming manner directly from a monocular RGB video?} Moving from multi-frame windows to frame-by-frame streaming poses two challenges. First, capturing temporal context: unlike window-based trackers that jointly reason over multiple frames, a streaming tracker observes only one new frame per step. Without history, each update reduces to independently matching the query point in the current frame, producing jittery trajectories. This calls for a bounded-size memory mechanism that carries context across frames. Second, 3D reasoning over noisy online geometry: depth and poses estimated online from monocular frames are inevitably noisy, and strictly causal inference cannot exploit future observations to correct these estimates. Since 3D updates are computed from correlations over the reconstructed point cloud, geometric noise directly biases them, and can push a track off its true image location onto a neighboring surface. Such errors compound in streaming, as subsequent updates are conditioned on the wrong local geometry.

To overcome these challenges, we present \textbf{StreamTAP3D}, a streaming 3D point tracker that processes each incoming frame upon arrival using only the current and past frames, with a per-frame cost independent of video length. To compensate for the multi-frame context that a single frame lacks, we introduce a hybrid long-short-term streaming memory: a lightweight first-in-first-out (FIFO) cache retains fine-grained motion tokens from recent frames, while a per-track Test-Time Training (TTT)~\cite{Sun2024LearningT} fast-weight memory compresses the entire history into constant-size states. To stabilize 3D refinement under noisy geometry, we propose a coupled 2D-3D streaming architecture. As 2D tracking on the image plane is immune to noise in the estimated geometry, we leverage pixel-accurate 2D predictions to guide 3D iterative refinement. Moreover, the two branches share visual features, letting the 3D branch benefit from abundant 2D tracking data despite scarce 3D annotations. Paired with a streaming reconstruction model, StreamTAP3D operates directly on raw monocular RGB streams without precomputed depth or poses.

Our main contributions are as follows:
\begin{itemize}
    \item We propose \textbf{StreamTAP3D}, a strictly causal 3D point tracker that operates on RGB streams with zero lookahead and a per-frame cost independent of video length.
    \item We introduce a hybrid long-short-term streaming memory with constant-size states, providing temporal context that improves tracking accuracy.
    \item We design a coupled 2D-3D streaming architecture that mitigates 3D drift under noisy online geometry and enables the 3D branch to benefit from abundant 2D data.
    \item At $384 \times 512$ on a single NVIDIA Hopper GPU, our full system, including reconstruction, tracks 256 points at 17 FPS under 10 GB VRAM, achieving accuracy competitive with methods that access future frames.
\end{itemize}

\begin{figure*}[t!]
    \centering
    \includegraphics[width=\linewidth]{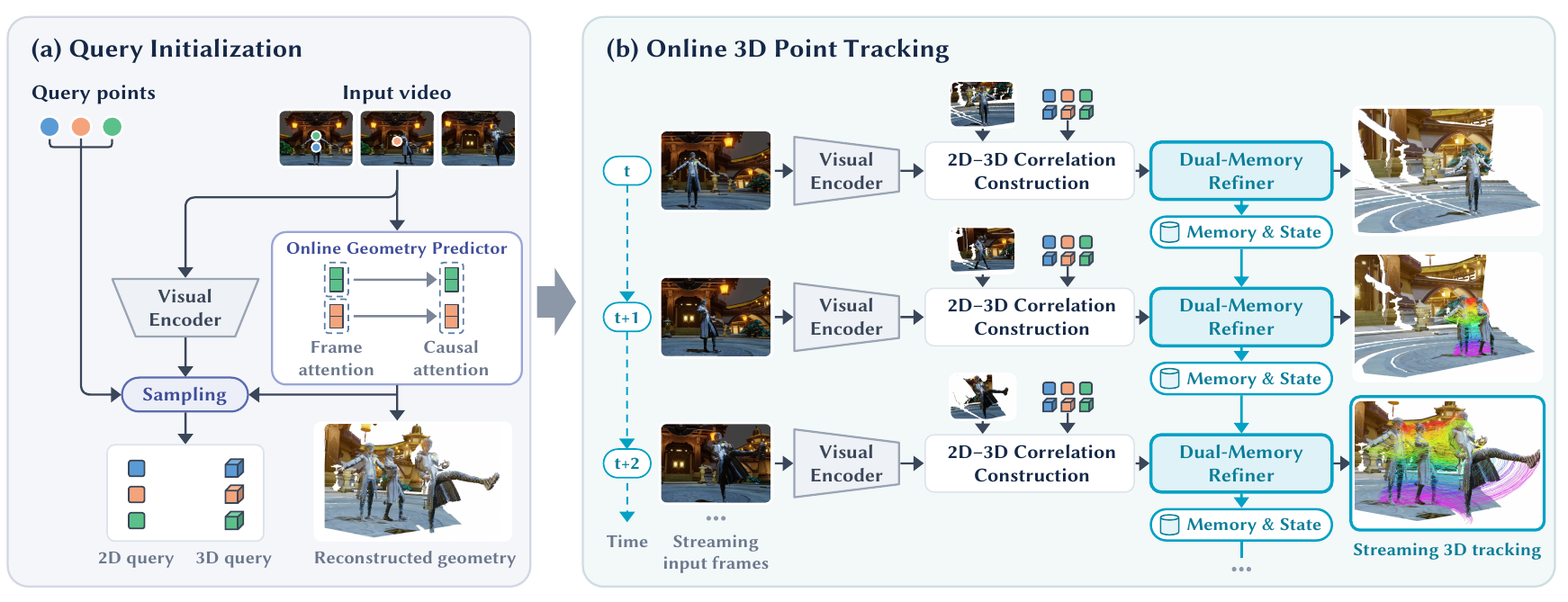}
    \vspace{-.2in}
    \caption{\small \textbf{Method Overview.} StreamTAP3D tracks 3D points by processing each frame upon arrival. \textbf{Left:} An \emph{Online Geometry Predictor} estimates per-frame geometry, while a shared \emph{Visual Encoder} extracts features from which \emph{2D queries} (squares) and \emph{3D queries} (cubes) are sampled. \textbf{Right:} For each incoming frame, 2D and 3D correlations are computed on the current frame only, and a \emph{Dual-Memory Refiner} iteratively updates both trajectories while carrying compact \emph{Memory \& State} to the next frame.
    }
    \label{fig:figure2}
\end{figure*}

\section{Related Work}

\noindent\textbf{2D Point Tracking.}
Long-term 2D point tracking estimates pixel trajectories and occlusion across videos~\cite{Harley2022ParticleVR, Doersch2023TAPIRTA, Karaev2023CoTrackerII, Karaev2025CoTracker3SA, Harley2025AllTrackerED, Lai2026CoWTrackerTB}, mostly offline or with sliding windows. Recent streaming trackers~\cite{dong2025spot, Aydemir2025TrackOn2EO} process frames one at a time via memory banks or recurrent states, but remain on the image plane; lifting them to 3D with streaming reconstruction merely reads out depth at the tracked pixel, without 3D reasoning. We instead jointly perform 2D and world-space 3D tracking in one streaming framework.

\noindent\textbf{3D Point Tracking.}
Extending point tracking to 3D~\cite{Koppula2024TAPVid3DAB, cao2025reconstructing} has attracted growing interest. Feed-forward approaches~\cite{zhu2026motioncrafter, Luo20264RC4R, Jiang2026OmniXAA, Sucar2026VDPM4V, Cheng2026MoSE3LW, Karhade2025Any4DUF, Xiao2025SpatialTrackerV23P} operate \emph{offline} on full videos, which precludes live streaming. SpaTracker~\cite{Xiao2024SpatialTrackerTA}, SceneTracker~\cite{Wang2024SceneTrackerLS}, DELTA~\cite{Ngo2024DELTADE}, TAPIP3D~\cite{Zhang2025TAPIP3DTA}, Track4World~\cite{Lu2026Track4WorldFW}, and TrackEverything~\cite{jain2026trackeverythinglonghorizondense} instead refine trajectories over overlapping \emph{sliding windows}, but must buffer future frames, and many assume precomputed depth or camera poses. Among online methods, St4RTrack~\cite{Feng2025ST4RTrackS4}, DynaDUSt3R~\cite{Jin2024Stereo4DLH}, and ZeroMSF~\cite{Liang2025ZeroShotMS} predict pointmaps or scene flow between frame pairs without memory across frames, while DynOMo~\cite{Seidenschwarz2024DynOMoOP} relies on costly per-frame optimization of dynamic 3D Gaussians. In contrast, StreamTAP3D is feed-forward, processes each frame upon arrival with zero lookahead, and maintains a bounded-size memory.

\noindent\textbf{Feed-Forward 3D Reconstruction Models.}
Feed-forward 3D foundation models have advanced camera and geometry estimation from monocular videos. DUSt3R~\cite{Wang2023DUSt3RG3}, MonST3R~\cite{Zhang2024MonST3RAS}, Align3R~\cite{lu2025align3r}, and VGGT~\cite{Wang2025VGGTVG} directly regress dense pointmaps and camera poses. Streaming reconstruction models~\cite{Wang2025Continuous3P, Lan2025STream3RSS,Chen2026LingBotMapGC, han2026abotrecon, zou2026iggt4d} incrementally estimate per-frame depth and poses via spatial memory or recurrent states. However, they do not establish point correspondences over time. StreamTAP3D pairs such streaming reconstruction with our tracker for simultaneous real-time reconstruction and tracking.

\noindent\textbf{Test-Time Training.}
Test-Time Training (TTT)~\cite{Sun2019TestTimeTW, Sun2024LearningT} and fast-weight memories~\cite{Schlag2021LinearTA} update compact hidden weights at inference via self-supervised gradient steps (\eg, the linear delta rule), compressing arbitrarily long sequences into constant-size states. Beyond sequence modeling, TTT layers have been adopted in long-context video modeling and 3D vision~\cite{Chen2025TTT3R3R, Jin2026ZipMapLS, Zhang2025TestTimeTD} as bounded-size long-horizon memory. Unlike these scene-level memories, we maintain a per-track fast-weight memory that compresses each trajectory's long-horizon history, complemented by a short FIFO cache for recent motion.

\section{Method}
\label{sec:method}

\noindent\textbf{Problem Formulation.}
Given a monocular RGB video stream $\{\mathbf{I}_t \in \mathbb{R}^{H \times W \times 3}\}_{t=0,1,\dots}$ and $N$ query points $\{\mathbf{q}_i = (t_i, x_i, y_i)\}_{i=1}^N$, we aim to estimate, for every track $i$ and every frame $t \geq t_i$, its 3D world-space position $\mathbf{X}_t^i \in \mathbb{R}^3$, its 2D pixel position $\mathbf{u}_t^i \in \mathbb{R}^2$, and the corresponding visibilities $v_t^{i,\text{3D}}, v_t^{i,\text{2D}} \in [0, 1]$. We operate in a \emph{strictly causal streaming} setting: the predictions for frame $t$ must be produced as soon as $\mathbf{I}_t$ arrives, using only $\mathbf{I}_t$ and a bounded-size state summarizing $\mathbf{I}_{0:t-1}$ (overview in \cref{fig:figure2}).

\subsection{Streaming Pipeline}
\label{sec:pipeline}

\noindent\textbf{Online Geometry Prediction.}
We estimate scene geometry on the fly with an Online Geometry Predictor (\cref{fig:figure2}, left), instantiated with a fine-tuned Abot-Recon~\cite{han2026abotrecon}. It alternates \emph{Frame Attention} and \emph{Causal Attention}, the latter restricted to the $11$ most recent frames, keeping its per-frame cost constant. It predicts a depth map $\mathbf{D}_t \in \mathbb{R}^{H \times W}$, intrinsics $\mathbf{K}_t \in \mathbb{R}^{3 \times 3}$, and camera-to-world extrinsics $\mathbf{T}_t^{\text{c2w}} = (\mathbf{R}_t, \mathbf{o}_t) \in \text{SE}(3)$. Each pixel $(x, y)$ is then unprojected into a world-space point map
\begin{equation}
    \mathbf{P}_t(x, y) = \mathbf{T}_t^{\text{c2w}} \big( \mathbf{D}_t(x, y) \, \mathbf{K}_t^{-1} [x, y, 1]^\top \big).
\end{equation}

\noindent\textbf{Query Initialization.}
Each frame $\mathbf{I}_t$ is also encoded by a Visual Encoder, shared by the 2D and 3D branches (\cref{sec:coupled}), into $S$-level multi-scale feature maps $\{\mathbf{F}_t^{(s)}\}_{s=1}^S$. When a query point $\mathbf{q}_i$ is introduced at frame $t_i$, the Sampling module initializes two templates: the \emph{2D query} $\mathbf{f}_i^{\text{2D}}$, a local feature patch bilinearly sampled from each $\mathbf{F}_{t_i}^{(s)}$ around $(x_i, y_i)$; and the \emph{3D query}, obtained by lifting $(x_i, y_i)$ to its world position $\mathbf{X}_{t_i}^i$ and gathering its $J$ nearest neighbors from the featurized point cloud $(\mathbf{P}_{t_i}, \mathbf{F}_{t_i}^{(s)})$, together with their relative 3D offsets, as support tokens.

\noindent\textbf{Single-Frame 2D-3D Correlation.}
Correlations are computed on the current frame only. At refinement iteration $m \in \{1, \dots, M\}$ of frame $t$, let $\hat{\mathbf{u}}_t^{i,m-1}$ and $\hat{\mathbf{X}}_t^{i,m-1}$ be the current estimates of track $i$, initialized from the final predictions of frame $t-1$.
The \emph{2D correlation} $\mathbf{C}_t^{i,\text{2D}}$ contains the all-pairs inner products between the query patch $\mathbf{f}_i^{\text{2D}}$ and the local patch around $\hat{\mathbf{u}}_t^{i,m-1}$ at each scale.
For the \emph{3D correlation} $\mathbf{C}_t^{i,\text{3D}}$, we attach the multi-scale features to $\mathbf{P}_t$, retrieve the $J$ nearest neighbors of $\hat{\mathbf{X}}_t^{i,m-1}$ at each scale as context tokens, and apply bidirectional Neighborhood-to-Neighborhood (N2N) cross-attention~\cite{Zhang2025TAPIP3DTA} between them and the 3D query support tokens.

\noindent\textbf{Streaming State.}
After frame $t$ is processed, its dense feature maps and point cloud are discarded. The information carried to frame $t+1$ includes, for each track, its latest 2D/3D estimates, a short FIFO cache of compact per-frame tokens, and a constant-size fast-weight matrix (\cref{sec:memory}). Per-frame cost is thus independent of video length, growing only with the number of active tracks.

\begin{figure}[t]
    \centering
    \includegraphics[width=\linewidth]{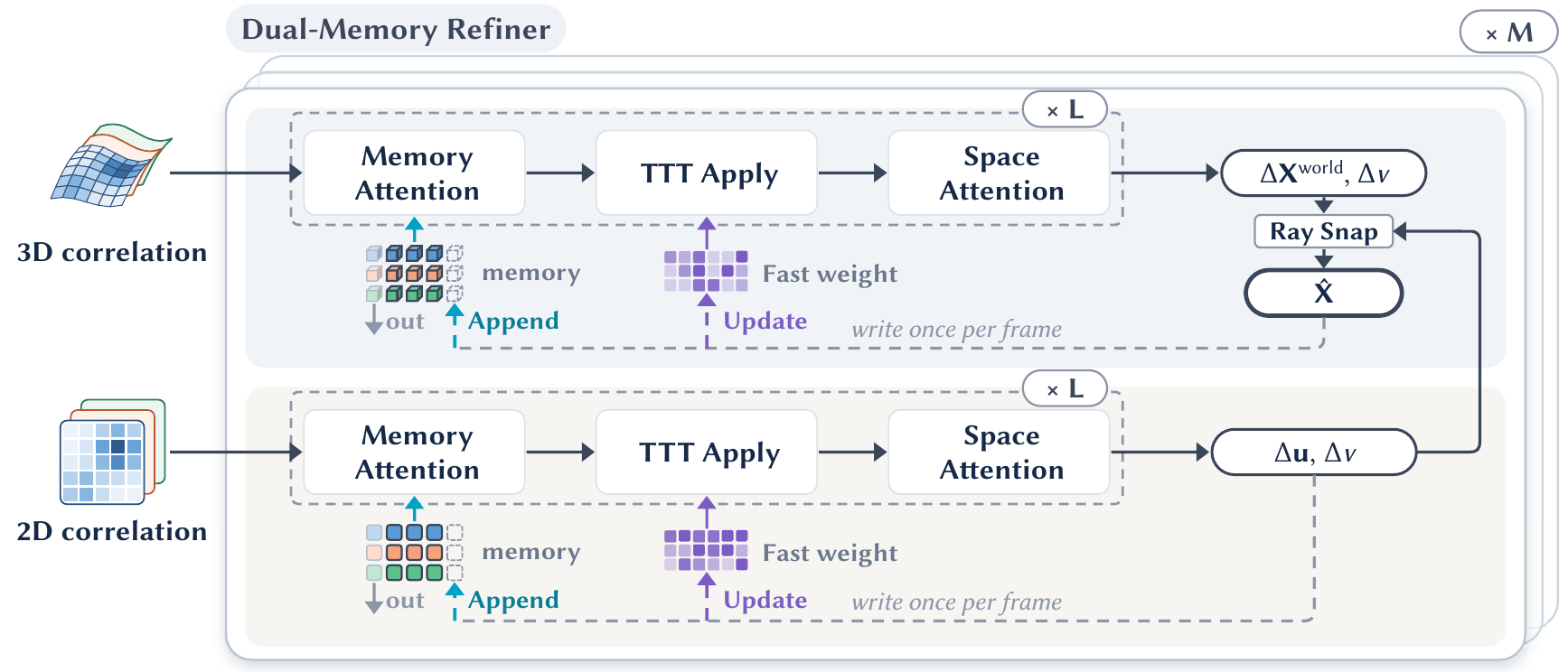}
    \vspace{-.2in}
    \caption{\textbf{Dual-Memory Refiner.} Parallel 2D and 3D branches iteratively refine track estimates; 3D updates are constrained by 2D-guided viewing rays (\emph{Ray Snap}). Both memories are written once per frame with the converged predictions.}
    \label{fig:figure3}
\end{figure}

\subsection{Hybrid Long-Short-Term Memory}
\label{sec:memory}

A single frame provides no motion context, so naive per-frame updates produce jittery trajectories. The Dual-Memory Refiner (\cref{fig:figure3}) supplies this context with two memories. Each of its 2D and 3D branches stacks $L$ layers of \emph{Memory Attention}, \emph{TTT Apply}, and cross-track \emph{Space Attention}, run for $M$ iterations per frame; both memories are read at every iteration but written once per frame.

\noindent\textbf{Short-Term FIFO Memory.}
Each branch keeps a per-track FIFO cache of capacity $B$. The entry for a past frame $\tau \in \{t-B, \dots, t-1\}$ stores only a compact feature $\mathbf{h}_\tau^i$, which encodes the converged correlation $\mathbf{C}_\tau^i$ and visibility $v_\tau^i$, together with the converged coordinate $\mathbf{p}_\tau^i$ ($\mathbf{u}_\tau^i$ in 2D, $\mathbf{X}_\tau^i$ in 3D). At iteration $m$ of frame $t$, we append the current state $(\mathbf{h}_t^{i,m-1}, \hat{\mathbf{p}}_t^{i,m-1})$ and encode each of the $B+1$ steps into a token $\mathbf{z}_\tau^i \in \mathbb{R}^C$:
\begin{equation}\label{eq:token}
    \mathbf{z}_\tau^i = \text{Proj}\Big(\big[\, \mathbf{h}_\tau^i \;\big\|\; \phi(\mathbf{p}_\tau^i - \mathbf{p}_{\tau-1}^i, \, \mathbf{p}_\tau^i - \mathbf{p}_{\tau+1}^i) \,\big]\Big) + \mathbf{e}_\tau^{\text{time}},
\end{equation}
where $\phi(\cdot)$ is a sinusoidal encoding of the backward and forward displacements (zero-padded at the boundaries), $\text{Proj}(\cdot)$ is a linear projection, and $\mathbf{e}_\tau^{\text{time}}$ is a fixed 1D sinusoidal temporal encoding. Memory Attention runs over $\{\mathbf{z}_\tau^i\}_{\tau=t-B}^{t}$; all $B+1$ tokens are propagated through the $L$ layers, and only the output at the current step $t$ is used to predict the update.

\noindent\textbf{Long-Term Fast-Weight Memory.}
To retain history beyond the FIFO window, we give each track a Test-Time Training (TTT)~\cite{Sun2024LearningT} fast-weight matrix $\mathbf{W}_t^i \in \mathbb{R}^{d \times d}$ in every layer of each branch, which compresses its entire history into a constant-size state. We omit layer and branch indices for brevity.
\emph{Read (TTT Apply).} At every iteration of frame $t$, each Memory Attention output $\mathbf{z}_\tau^i$ in the window is projected into a read vector $\mathbf{r}_\tau^i \in \mathbb{R}^d$, which retrieves long-horizon context from the previous frame's fast weights:
\begin{equation}
    \mathbf{a}_\tau^i = \mathbf{W}_{t-1}^i \mathbf{r}_\tau^i, \qquad \mathbf{z}_\tau^i \leftarrow \mathbf{z}_\tau^i + \text{Proj}(\mathbf{a}_\tau^i).
\end{equation}
\emph{Write (Update).} After the final iteration, we resample features at the converged prediction, including recomputed correlation and visibility, and encode them with a separate MLP into a write token $\hat{\mathbf{z}}_t^i$ per track, shared by all layers%
. Each layer separately projects $\hat{\mathbf{z}}_t^i$ into a key $\mathbf{k}_t^i \in \mathbb{R}^d$, a target $\mathbf{y}_t^i \in \mathbb{R}^d$, and a step size $\beta_t^i \in (0, 1)$. One gradient step on the self-supervised objective $\mathcal{L}_{\text{TTT}}(\mathbf{W}) = \frac{1}{2}\|\mathbf{W}\mathbf{k}_t^i - \mathbf{y}_t^i\|_2^2$ gives the linear delta-rule update
\begin{equation}\label{eq:ttt-write}
    \mathbf{W}_t^i = \mathbf{W}_{t-1}^i + \beta_t^i \big( \mathbf{y}_t^i - \mathbf{W}_{t-1}^i \mathbf{k}_t^i \big) (\mathbf{k}_t^i)^\top.
\end{equation}

\noindent\textbf{Space Attention and Outputs.}
Space Attention then exchanges information among all $N$ tracks at each time step in the window, through a small set of learnable virtual tracks~\cite{Karaev2023CoTrackerII}. After the $L$ layers, the 2D branch predicts $(\Delta \mathbf{u}_t^i, \Delta v_t^{i,\text{2D}})$ and the 3D branch predicts $(\Delta \mathbf{X}_t^{i}, \Delta v_t^{i,\text{3D}})$. The 2D update is applied directly, while the 3D update is first constrained by the ray snapping described next.

\begin{table*}[tb!]
\centering
\captionsetup{skip=2pt}
\caption{\textbf{Quantitative Results on World-Coordinate Tracking.} We report the $\text{APD}_{\text{3D}}$ and $\text{AJ}_{\text{3D}}$ over the first 64 frames on the four WorldTrack subsets.
Non-causal methods (\textcolor{refgray}{gray}), which access future frames, are listed for reference only; we highlight the \besthl{best} and \secondbesthl{second} best results among causal streaming trackers.}
\label{tab:worldcoord-tracking}
\setlength{\tabcolsep}{4pt}
\def\groupgap{12pt}
\renewcommand{\arraystretch}{0.95}
\resizebox{\textwidth}{!}{%
    \begin{tabular}{l*{5}{>{\hspace{\groupgap}}cc}}
    \toprule
    \multirow{2}{*}{\textbf{Method}}
    & \multicolumn{2}{>{\hspace{\groupgap}}c}{\textbf{ADT}}
    & \multicolumn{2}{>{\hspace{\groupgap}}c}{\textbf{Dynamic Replica}}
    & \multicolumn{2}{>{\hspace{\groupgap}}c}{\textbf{PointOdyssey}}
    & \multicolumn{2}{>{\hspace{\groupgap}}c}{\textbf{PStudio}}
    & \multicolumn{2}{>{\hspace{\groupgap}}c}{\textbf{Avg.}} \\
    \cmidrule(l{\groupgap}r){2-3}
    \cmidrule(l{\groupgap}r){4-5}
    \cmidrule(l{\groupgap}r){6-7}
    \cmidrule(l{\groupgap}r){8-9}
    \cmidrule(l{\groupgap}r){10-11}
    & $\text{APD}_{\text{3D}}\uparrow$ & $\text{AJ}_{\text{3D}}\uparrow$
    & $\text{APD}_{\text{3D}}\uparrow$ & $\text{AJ}_{\text{3D}}\uparrow$
    & $\text{APD}_{\text{3D}}\uparrow$ & $\text{AJ}_{\text{3D}}\uparrow$
    & $\text{APD}_{\text{3D}}\uparrow$ & $\text{AJ}_{\text{3D}}\uparrow$
    & $\text{APD}_{\text{3D}}\uparrow$ & $\text{AJ}_{\text{3D}}\uparrow$ \\
    \midrule
    \multicolumn{11}{l}{\textit{\textcolor{refgray}{Offline Full-Sequence Trackers}}} \\
    \rg{TraceAnything~\citep{Liu2025TraceAR}} & \rg{58.93} & \rg{46.88} & \rg{45.39} & \rg{36.49} & \rg{39.76} & \rg{23.85} & \rg{49.78} & \rg{38.35} & \rg{48.47} & \rg{36.39} \\
    \rg{Any4D~\citep{Karhade2025Any4DUF}} & \rg{57.91} & \rg{45.86} & \rg{54.55} & \rg{45.23} & \rg{50.76} & \rg{31.52} & \rg{53.76} & \rg{41.47} & \rg{54.25} & \rg{41.02} \\
    \rg{SpaTrackerV2~\citep{Xiao2025SpatialTrackerV23P}} & \rg{67.23} & \rg{59.71} & \rg{54.47} & \rg{45.96} & \rg{49.49} & \rg{33.52} & \rg{58.56} & \rg{46.64} & \rg{57.44} & \rg{46.46} \\
    \rg{V-DPM~\citep{Sucar2026VDPM4V}} & \rg{56.22} & \rg{44.82} & \rg{54.56} & \rg{45.11} & \rg{54.34} & \rg{33.95} & \rg{59.76} & \rg{46.55} & \rg{56.22} & \rg{42.61} \\
    \rg{OpenD4RT~\citep{Zhang2025EfficientlyRD}} & \rg{56.45} & \rg{48.03} & \rg{53.95} & \rg{45.21} & \rg{50.67} & \rg{34.32} & \rg{60.77} & \rg{50.50} & \rg{55.46} & \rg{44.52} \\
    \rg{Point4D~\citep{Jeon2026Point4DL4}} & \rg{51.20} & \rg{39.17} & \rg{55.37} & \rg{46.51} & \rg{53.77} & \rg{34.31} & \rg{59.99} & \rg{48.24} & \rg{55.08} & \rg{42.06} \\
    \rg{4RC~\citep{Luo20264RC4R}} & \rg{65.90} & \rg{53.12} & \rg{59.96} & \rg{50.09} & \rg{59.69} & \rg{37.63} & \rg{60.17} & \rg{46.82} & \rg{61.43} & \rg{46.92} \\
    \midrule
    \multicolumn{11}{l}{\textit{\textcolor{refgray}{Sliding-Window Trackers}}} \\
    \rg{SpaTracker~\citep{Xiao2024SpatialTrackerTA}} & \rg{47.30} & \rg{41.38} & \rg{36.06} & \rg{28.53} & \rg{37.62} & \rg{24.58} & \rg{52.45} & \rg{40.94} & \rg{43.36} & \rg{33.86} \\
    \rg{DELTA~\citep{Ngo2024DELTADE}} & \rg{57.27} & \rg{51.11} & \rg{52.80} & \rg{43.67} & \rg{47.58} & \rg{32.33} & \rg{58.13} & \rg{44.07} & \rg{53.95} & \rg{42.80} \\
    \rg{Track4World~\citep{Lu2026Track4WorldFW}} & \rg{59.78} & \rg{54.07} & \rg{53.08} & \rg{45.29} & \rg{50.36} & \rg{34.67} & \rg{56.41} & \rg{44.34} & \rg{54.91} & \rg{44.59} \\
    \rg{TAPIP3D~\citep{Zhang2025TAPIP3DTA}} & \rg{53.72} & \rg{47.66} & \rg{54.82} & \rg{47.23} & \rg{54.34} & \rg{38.64} & \rg{57.77} & \rg{47.69} & \rg{55.16} & \rg{45.30} \\
    \midrule
    \multicolumn{11}{l}{\textit{Causal Streaming Trackers}} \\
    St4RTrack~\citep{Feng2025ST4RTrackS4} & 44.45 & 34.93 & 47.20 & 38.51 & 37.55 & 22.66 & 40.92 & 31.82 & 42.53 & 31.98 \\
    DynaDUSt3R~\citep{Jin2024Stereo4DLH} & 47.16 & 36.88 & 41.01 & 33.38 & 38.36 & 22.94 & 50.26 & 38.56 & 44.20 & 32.94 \\
    ZeroMSF~\citep{Liang2025ZeroShotMS} & 44.17 & 34.98 & 46.93 & 38.90 & 35.58 & 21.53 & 41.69 & 32.27 & 42.09 & 31.92 \\
    SPOT~\citep{dong2025spot} + Abot-Recon & 55.96 & \secondbesthl{50.44} & \secondbesthl{57.06} & \secondbesthl{48.51} & \secondbesthl{52.29} & 35.95 & 57.25 & 45.92 & \secondbesthl{55.64} & 45.21 \\
    Track-On2~\citep{Aydemir2025TrackOn2EO} + Abot-Recon & \secondbesthl{56.06} & \besthl{51.10} & 56.39 & 47.82 & 51.84 & \secondbesthl{36.18} & \secondbesthl{58.18} & \secondbesthl{47.88} & 55.62 & \secondbesthl{45.75} \\
    \oursrow Ours & \besthl{56.44} & 50.07 & \besthl{59.38} & \besthl{50.72} & \besthl{57.63} & \besthl{40.87} & \besthl{59.35} & \besthl{48.57} & \besthl{58.20} & \besthl{47.56} \\
    \bottomrule
    \end{tabular}%
 }
\vspace{-6pt}
\end{table*}

\subsection{Coupled 2D-3D Streaming Architecture}
\label{sec:coupled}
3D correlations, computed on the reconstructed point cloud, are biased by streaming reconstruction noise, whereas 2D matching on the image plane is geometry-independent. We thus couple the two branches in two ways. First, they share the Visual Encoder, so that the 3D branch can benefit from abundant 2D tracking data. Second, 2D predictions guide 3D refinement via ray snapping.

\noindent\textbf{2D-Guided Ray Snapping.}
Once a foreground point strays onto the background, its nearest 3D neighbors are dominated by background points, which hinders recovery in later iterations. We therefore constrain the 3D update with the 2D prediction at every iteration. The updated 2D position $\hat{\mathbf{u}}_t^{i,m} = (\hat{x}_t^{i,m}, \hat{y}_t^{i,m})$ defines a world-space ray from the camera center $\mathbf{o}_t$ with direction $\boldsymbol{\ell}_t^i = \mathbf{R}_t \mathbf{K}_t^{-1} [\hat{x}_t^{i,m}, \hat{y}_t^{i,m}, 1]^\top$. Let $\mathbf{n}_t = \mathbf{R}_t [0, 0, 1]^\top$ be the camera principal axis and $z_t^{i,m-1} = \langle \hat{\mathbf{X}}_t^{i,m-1} - \mathbf{o}_t, \mathbf{n}_t \rangle$ the depth of the current estimate. We apply the 3D update along the 2D-guided viewing ray:
\begin{equation}
    \hat{\mathbf{X}}_t^{i,m} = \mathbf{o}_t + \big(z_t^{i,m-1} + \langle \Delta \mathbf{X}_t^{i}, \mathbf{n}_t \rangle\big)\, \boldsymbol{\ell}_t^i.
\end{equation}

\subsection{Training}
\label{sec:training}

The Online Geometry Predictor is fine-tuned separately, while the tracker is trained on noise-augmented ground-truth depth and camera parameters, replaced by the predictor's outputs at inference. To make the 3D branch agnostic to the camera trajectory, we further map each training clip into a random world frame by applying a randomly sampled smooth sequence of rigid transformations to its camera-frame points, which is ablated in \cref{tab:ablation-spatial}. The tracker is trained on \(T\)-frame clips, unrolled frame by frame as at inference. Each branch is supervised if the corresponding annotation is available, with exponentially increasing weights assigned to successive refinement iterations ($\eta = 0.8$):
\begin{equation}\label{eq:loss}
    \mathcal{L} = \sum_{t=0}^{T-1} \sum_{m=1}^{M} \eta^{M-m} \left( \lambda_{\text{traj}} \mathcal{L}_{\text{traj}}^{(t,m)} + \lambda_{\text{bce}} \mathcal{L}_{\text{bce}}^{(t,m)} \right),
\end{equation}
where $\mathcal{L}_{\text{traj}}^{(t,m)}$ combines camera-frame 3D Euclidean errors of both the pre-snapping estimate $\hat{\mathbf{X}}_t^{i,m-1} + \Delta \mathbf{X}_t^i$ and the snapped estimate $\hat{\mathbf{X}}_t^{i,m}$, scaled by $f / z^{\text{gt}}$ (focal length over ground-truth depth), with an $L_1$ loss on 2D trajectories, and $\mathcal{L}_{\text{bce}}^{(t,m)}$ is a visibility BCE loss for both branches. We train on a mixture of 2D and 3D tracking datasets (\cref{sec:exp_setting}); for samples with only 2D annotations, the 3D branch is skipped and only the 2D terms are applied.

\section{Experiments}
\label{sec:exp}

\subsection{Experimental Setting}
\label{sec:exp_setting}

\begin{table*}[tb!]
\centering
\captionsetup{skip=2pt}
\caption{\textbf{Quantitative Results on Long-Sequence World-Coordinate Tracking.} We report the $\text{APD}_{\text{3D}}$ and $\text{AJ}_{\text{3D}}$ evaluated on full-length sequences across the four WorldTrack subsets.
Non-causal methods (\textcolor{refgray}{gray}), which access future frames, are listed for reference only; we highlight the \besthl{best} and \secondbesthl{second} best results among causal streaming trackers.}
\label{tab:worldcoord-tracking-full}
\setlength{\tabcolsep}{4pt}
\def\groupgap{12pt}
\renewcommand{\arraystretch}{0.95}
\resizebox{\textwidth}{!}{%
    \begin{tabular}{l*{5}{>{\hspace{\groupgap}}cc}}
    \toprule
    \multirow{2}{*}{\textbf{Method}}
    & \multicolumn{2}{>{\hspace{\groupgap}}c}{\textbf{ADT}}
    & \multicolumn{2}{>{\hspace{\groupgap}}c}{\textbf{Dynamic Replica}}
    & \multicolumn{2}{>{\hspace{\groupgap}}c}{\textbf{PointOdyssey}}
    & \multicolumn{2}{>{\hspace{\groupgap}}c}{\textbf{PStudio}}
    & \multicolumn{2}{>{\hspace{\groupgap}}c}{\textbf{Avg.}} \\
    \cmidrule(l{\groupgap}r){2-3}
    \cmidrule(l{\groupgap}r){4-5}
    \cmidrule(l{\groupgap}r){6-7}
    \cmidrule(l{\groupgap}r){8-9}
    \cmidrule(l{\groupgap}r){10-11}
    & $\text{APD}_{\text{3D}}\uparrow$ & $\text{AJ}_{\text{3D}}\uparrow$
    & $\text{APD}_{\text{3D}}\uparrow$ & $\text{AJ}_{\text{3D}}\uparrow$
    & $\text{APD}_{\text{3D}}\uparrow$ & $\text{AJ}_{\text{3D}}\uparrow$
    & $\text{APD}_{\text{3D}}\uparrow$ & $\text{AJ}_{\text{3D}}\uparrow$
    & $\text{APD}_{\text{3D}}\uparrow$ & $\text{AJ}_{\text{3D}}\uparrow$ \\
    \midrule
    \multicolumn{11}{l}{\textit{\textcolor{refgray}{Offline \& Sliding-Window Trackers}}} \\
    \rg{Point4D~\citep{Jeon2026Point4DL4}} & \rg{44.04} & \rg{26.12} & \rg{54.63} & \rg{44.72} & \rg{52.67} & \rg{32.15} & \rg{56.62} & \rg{42.90} & \rg{51.99} & \rg{36.47} \\
    \rg{SpaTracker~\citep{Xiao2024SpatialTrackerTA}} & \rg{32.20} & \rg{23.06} & \rg{35.86} & \rg{28.18} & \rg{34.72} & \rg{21.79} & \rg{48.26} & \rg{35.04} & \rg{37.76} & \rg{27.02} \\
    \rg{DELTA~\citep{Ngo2024DELTADE}} & \rg{38.14} & \rg{27.44} & \rg{51.78} & \rg{42.32} & \rg{45.33} & \rg{29.75} & \rg{53.60} & \rg{37.64} & \rg{47.21} & \rg{34.29} \\
    \rg{Track4World~\citep{Lu2026Track4WorldFW}} & \rg{33.96} & \rg{25.78} & \rg{35.58} & \rg{28.90} & \rg{35.66} & \rg{23.05} & \rg{47.54} & \rg{33.96} & \rg{38.19} & \rg{27.92} \\
    \rg{TAPIP3D~\citep{Zhang2025TAPIP3DTA}} & \rg{44.04} & \rg{35.20} & \rg{52.51} & \rg{44.59} & \rg{50.30} & \rg{34.18} & \rg{53.50} & \rg{41.61} & \rg{50.09} & \rg{38.89} \\
    \midrule
    \multicolumn{11}{l}{\textit{Causal Streaming Trackers}} \\
    St4RTrack~\citep{Feng2025ST4RTrackS4} & 36.18 & 19.38 & 47.37 & 37.36 & 37.05 & 20.64 & 39.38 & 28.91 & 40.00 & 26.57 \\
    DynaDUSt3R~\citep{Jin2024Stereo4DLH} & 38.35 & 20.77 & 40.69 & 32.14 & 37.90 & 21.07 & 47.26 & 34.61 & 41.05 & 27.15 \\
    ZeroMSF~\citep{Liang2025ZeroShotMS} & 34.08 & 18.51 & 44.86 & 36.12 & 33.69 & 19.17 & 39.77 & 29.14 & 38.10 & 25.74 \\
    SPOT~\citep{dong2025spot} + Abot-Recon & 35.78 & 25.73 & 54.73 & 45.76 & 48.35 & 31.52 & 52.82 & 39.36 & 47.92 & 35.59 \\
    Track-On2~\citep{Aydemir2025TrackOn2EO} + Abot-Recon & \besthl{42.36} & \besthl{37.99} & \secondbesthl{54.94} & \secondbesthl{46.06} & \secondbesthl{49.28} & \secondbesthl{33.46} & \secondbesthl{54.65} & \besthl{43.41} & \secondbesthl{50.31} & \besthl{40.23} \\
    \oursrow Ours & \secondbesthl{42.29} & \secondbesthl{30.84} & \besthl{58.28} & \besthl{49.14} & \besthl{54.47} & \besthl{37.31} & \besthl{55.89} & \secondbesthl{42.57} & \besthl{52.73} & \secondbesthl{39.97} \\
    \bottomrule
    \end{tabular}%
}
\vspace{-6pt}
\end{table*}

\begin{figure*}[t!]
    \centering
    \includegraphics[width=\linewidth]{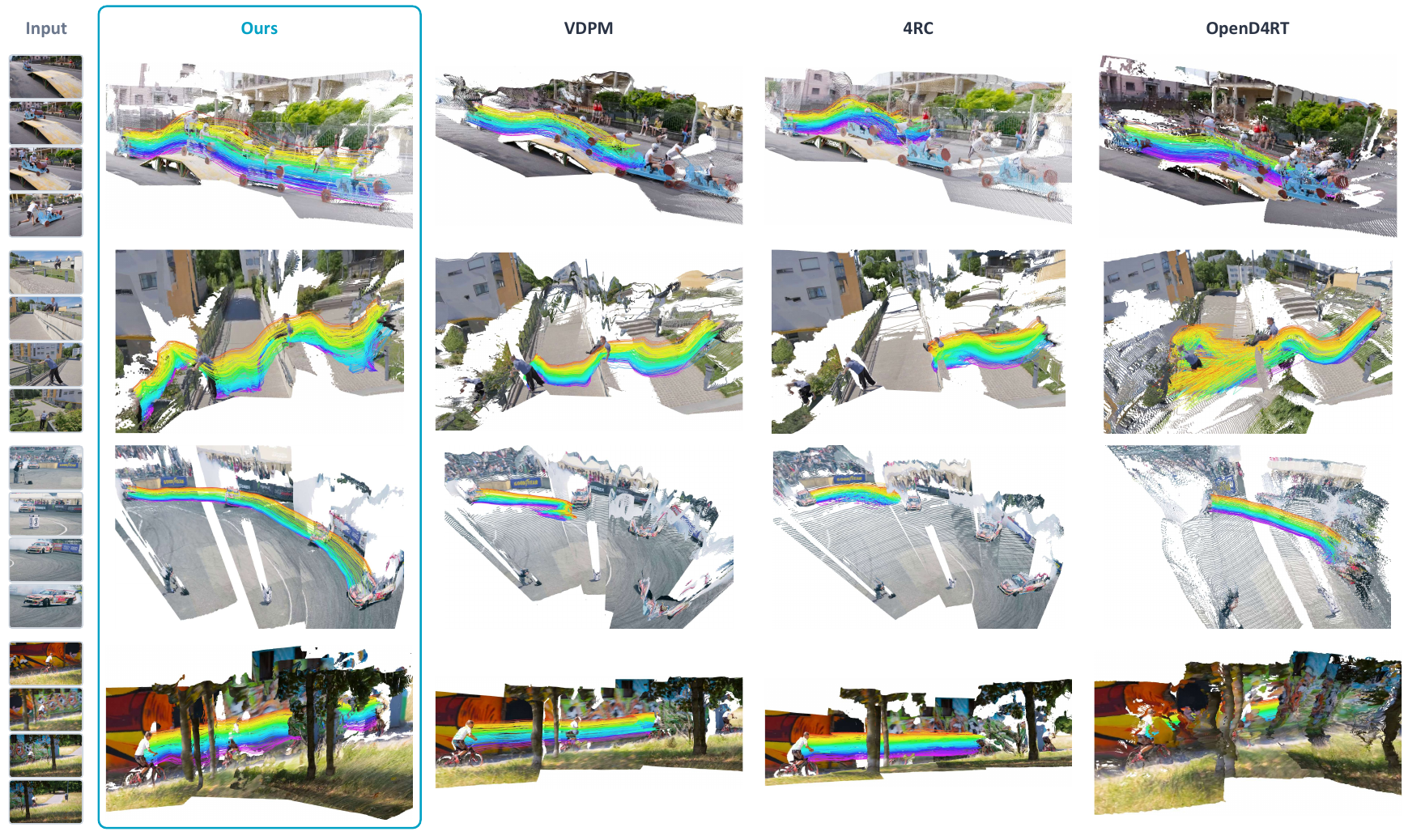}
    \vspace{-.2in}
    \caption{\small \textbf{Qualitative Results of 3D Tracking across Different Methods.} We visualize world-coordinate 3D point trajectories on challenging sequences and we colorize the dynamic foreground tracklines.
    }
    \label{fig:figure4}
\end{figure*}

\noindent\textbf{Training Datasets.}
We train StreamTAP3D end-to-end on a mixture of 2D and 3D point tracking datasets, including Kubric~\cite{Greff2022KubricAS}, DynamicReplica~\cite{Karaev2023DynamicStereoCD}, PointOdyssey~\cite{Zheng2023PointOdysseyAL}, Stereo4D~\cite{Jin2024Stereo4DLH}, FlyingThings3D~\cite{Mayer2015ALD}, Driving~\cite{Mayer2015ALD}, Monkaa~\cite{Mayer2015ALD} and Spring~\cite{Mehl2023SpringAH}. Furthermore, to alleviate the scarcity of diverse 3D point tracking data, we build a custom synthetic 3D dataset of approximately 170k frames rendered with Blender, specifically curated to feature large camera motions and complex 3D motion patterns. We also fine-tune Abot-Recon as our Online Geometry Predictor (details in the supplementary material). For all datasets, we use only the training splits.

\noindent\textbf{Implementation Details.}
To progressively adapt the streaming memory to long-horizon dynamics, we adopt a three-stage sequence-length curriculum, training on 24-, 48-, and 64-frame sequences for 50k, 10k, and 10k iterations, respectively. The shared Visual Encoder is a CNN feature pyramid with base stride $s=4$ and $S=4$ levels. The Dual-Memory Refiner uses $L=3$ transformer layers, $M=4$ refinement iterations per frame, and a FIFO capacity of $B=15$. We train at $384 \times 512$ using AdamW with a learning rate of $1 \times 10^{-4}$ and a weight decay of $5 \times 10^{-4}$.

\noindent\textbf{Benchmarks and Metrics.}
We evaluate on WorldTrack~\cite{Feng2025ST4RTrackS4}, which covers ADT~\cite{Koppula2024TAPVid3DAB}, DynamicReplica ({DR})~\cite{Karaev2023DynamicStereoCD}, PointOdyssey ({PO})~\cite{Zheng2023PointOdysseyAL}, and Panoptic Studio ({PStudio})~\cite{Koppula2024TAPVid3DAB}. Following the protocol of Track4World~\cite{Lu2026Track4WorldFW}, we report $\text{APD}_{\text{3D}}$, the average fraction of points within a set of world-space distance thresholds after global scale-and-shift alignment, and $\text{AJ}_{\text{3D}}$, which jointly measures position and occlusion accuracy. All methods are re-evaluated with the same protocol.

\begin{table*}[tb!]
\centering
\captionsetup{skip=2pt}
\caption{\textbf{Quantitative Results on 2D Point Tracking.} We report the standard Average Position Accuracy ($\delta_{\text{avg}}\uparrow$), Average Jaccard (AJ$\uparrow$), and Occlusion Accuracy (OA$\uparrow$) across four 2D point tracking benchmarks ({DAVIS}, {DriveTrack}, {Kinetics}, and {RoboTAP}) and their average (Avg.).
Non-causal methods (\textcolor{refgray}{gray}), which access future frames, are listed for reference only; we highlight the \besthl{best} and \secondbesthl{second} best results among causal streaming trackers.}
\label{tab:2d-point-tracking}
\setlength{\tabcolsep}{4pt}
\def\groupgap{22.4pt}
\renewcommand{\arraystretch}{0.95}
\resizebox{\textwidth}{!}{%
    \begin{tabular}{l*{5}{>{\hspace{\groupgap}}ccc}}
    \toprule
    \multirow{2}{*}{\textbf{Method}}
    & \multicolumn{3}{>{\hspace{\groupgap}}c}{\textbf{DAVIS}}
    & \multicolumn{3}{>{\hspace{\groupgap}}c}{\textbf{DriveTrack}}
    & \multicolumn{3}{>{\hspace{\groupgap}}c}{\textbf{Kinetics}}
    & \multicolumn{3}{>{\hspace{\groupgap}}c}{\textbf{RoboTAP}}
    & \multicolumn{3}{>{\hspace{\groupgap}}c}{\textbf{Avg.}} \\
    \cmidrule(l{\groupgap}r){2-4}
    \cmidrule(l{\groupgap}r){5-7}
    \cmidrule(l{\groupgap}r){8-10}
    \cmidrule(l{\groupgap}r){11-13}
    \cmidrule(l{\groupgap}r){14-16}
    & $\delta_{\text{avg}}\uparrow$ & {AJ}$\uparrow$ & {OA}$\uparrow$
    & $\delta_{\text{avg}}\uparrow$ & {AJ}$\uparrow$ & {OA}$\uparrow$
    & $\delta_{\text{avg}}\uparrow$ & {AJ}$\uparrow$ & {OA}$\uparrow$
    & $\delta_{\text{avg}}\uparrow$ & {AJ}$\uparrow$ & {OA}$\uparrow$
    & $\delta_{\text{avg}}\uparrow$ & {AJ}$\uparrow$ & {OA}$\uparrow$ \\
    \midrule
    \multicolumn{16}{l}{\textit{\textcolor{refgray}{Offline \& Sliding-Window Trackers}}} \\
    \rg{TAPIR~\citep{Doersch2023TAPIRTA}} & \rg{70.7} & \rg{58.5} & \rg{87.3} & \rg{57.7} & \rg{45.3} & \rg{77.3} & \rg{65.6} & \rg{52.3} & \rg{86.9} & \rg{71.7} & \rg{58.0} & \rg{86.4} & \rg{66.4} & \rg{53.5} & \rg{84.5} \\
    \rg{BootsTAPIR~\cite{Doersch2024BootsTAPBT}} & \rg{74.4} & \rg{62.1} & \rg{89.6} & \rg{62.4} & \rg{48.8} & \rg{78.3} & \rg{69.5} & \rg{56.5} & \rg{88.7} & \rg{77.6} & \rg{63.4} & \rg{88.5} & \rg{71.0} & \rg{57.7} & \rg{86.3} \\
    \rg{LocoTrack~\cite{Cho2024LocalAC}} & \rg{75.2} & \rg{63.0} & \rg{87.3} & \rg{61.5} & \rg{48.2} & \rg{78.4} & \rg{70.4} & \rg{56.0} & \rg{87.5} & \rg{78.7} & \rg{65.2} & \rg{88.6} & \rg{71.5} & \rg{58.1} & \rg{85.5} \\
    \rg{PIPs++~\cite{Zheng2023PointOdysseyAL}} & \rg{69.4} & \rg{46.5} & \rg{75.2} & \rg{52.0} & \rg{37.6} & \rg{87.0} & \rg{66.6} & \rg{41.0} & \rg{67.5} & \rg{73.6} & \rg{48.5} & \rg{70.4} & \rg{65.4} & \rg{43.4} & \rg{75.0} \\
    \rg{DELTA~\citep{Ngo2024DELTADE}} & \rg{76.4} & \rg{62.9} & \rg{88.7} & \rg{67.6} & \rg{53.2} & \rg{80.1} & \rg{66.4} & \rg{51.9} & \rg{84.4} & \rg{73.4} & \rg{62.7} & \rg{87.7} & \rg{71.0} & \rg{57.7} & \rg{85.2} \\
    \rg{Track4World~\citep{Lu2026Track4WorldFW}} & \rg{74.5} & \rg{61.4} & \rg{86.0} & \rg{60.6} & \rg{45.6} & \rg{68.6} & \rg{71.5} & \rg{59.3} & \rg{90.5} & \rg{81.9} & \rg{71.0} & \rg{93.4} & \rg{72.1} & \rg{59.3} & \rg{84.6} \\
    \rg{CoTracker3~\citep{Karaev2025CoTracker3SA}} & \rg{77.4} & \rg{64.8} & \rg{90.4} & \rg{65.9} & \rg{52.9} & \rg{84.6} & \rg{71.9} & \rg{58.6} & \rg{89.8} & \rg{80.8} & \rg{68.2} & \rg{91.5} & \rg{74.0} & \rg{61.1} & \rg{89.1} \\
    \rg{AllTracker~\citep{Harley2025AllTrackerED}} & \rg{76.3} & \rg{63.3} & \rg{90.1} & \rg{63.1} & \rg{50.4} & \rg{79.6} & \rg{72.3} & \rg{59.2} & \rg{90.4} & \rg{82.8} & \rg{71.1} & \rg{92.9} & \rg{73.6} & \rg{61.0} & \rg{88.3} \\
    \midrule
    \multicolumn{16}{l}{\textit{Causal Streaming Trackers}} \\
    SPOT~\citep{dong2025spot} & 74.9 & 60.4 & 86.6 & 60.7 & 46.9 & 79.0 & 65.3 & 50.3 & 81.7 & 70.0 & 55.7 & 81.5 & 67.7 & 53.3 & 82.2 \\
    Track-On2~\citep{Aydemir2025TrackOn2EO} & \besthl{79.4} & \besthl{66.3} & \besthl{91.4} & \besthl{71.0} & \besthl{57.2} & \besthl{88.8} & \secondbesthl{71.9} & \secondbesthl{57.5} & \besthl{90.4} & \secondbesthl{82.1} & \secondbesthl{69.7} & \secondbesthl{93.4} & \secondbesthl{76.1} & \besthl{62.7} & \besthl{91.0} \\
    \oursrow Ours & \secondbesthl{78.2} & \secondbesthl{62.8} & \secondbesthl{89.2} & \secondbesthl{70.6} & \secondbesthl{55.2} & \secondbesthl{85.3} & \besthl{72.2} & \besthl{58.2} & \secondbesthl{89.3} & \besthl{84.0} & \besthl{71.8} & \besthl{93.7} & \besthl{76.3} & \secondbesthl{62.0} & \secondbesthl{89.4} \\
    \bottomrule
    \end{tabular}%
}
\vspace{-6pt}
\end{table*}

\noindent\textbf{Baselines.}
We compare against offline full-sequence, sliding-window, and causal streaming 3D trackers (\cref{tab:worldcoord-tracking,tab:worldcoord-tracking-full}). The causal baselines include the pairwise feed-forward methods St4RTrack~\cite{Feng2025ST4RTrackS4}, DynaDUSt3R~\cite{Jin2024Stereo4DLH}, and ZeroMSF~\cite{Liang2025ZeroShotMS}, and the 2D streaming trackers SPOT~\cite{dong2025spot} and Track-On2~\cite{Aydemir2025TrackOn2EO} lifted with our fine-tuned Abot-Recon~\cite{han2026abotrecon}; for TAPIP3D~\cite{Zhang2025TAPIP3DTA}, we likewise replace its slow optimization-based MegaSaM geometry with Abot-Recon. 2D baselines are listed in \cref{tab:2d-point-tracking}.

\subsection{Point Tracking Evaluation}
\label{sec:point_tracking_eval}

\noindent\textbf{3D Point Tracking.}
\Cref{tab:worldcoord-tracking} compares 3D point tracking on WorldTrack, where the lifted 2D baselines share our Abot-Recon~\cite{han2026abotrecon} geometry. StreamTAP3D outperforms all causal streaming baselines in $\text{APD}_{\text{3D}}$ on every subset, especially on the highly dynamic DR and PO. Despite having no access to future frames, it achieves the best average $\text{AJ}_{\text{3D}}$ among all methods and is second only to the offline 4RC~\cite{Luo20264RC4R} in average $\text{APD}_{\text{3D}}$. \Cref{fig:figure4} further shows smoother and more temporally coherent world-space trajectories than competing methods.

\noindent\textbf{Long-Sequence 3D Point Tracking.}
We further evaluate on the full-length WorldTrack sequences (128--300 frames) in \cref{tab:worldcoord-tracking-full}. Since most offline methods run out of memory at these lengths, we only compare with sliding-window trackers, Point4D, and causal streaming trackers. Our method achieves the best average $\text{APD}_{\text{3D}}$. On ADT, where our positional accuracy is similar to that of Track-On2, our lower $\text{AJ}_{\text{3D}}$ stems from visibility: the visibility predicted by our 3D branch is less accurate than that of the lifted Track-On2 over long sequences.

\noindent\textbf{2D Point Tracking.}
 We also evaluate our 2D tracking branch on four 2D benchmarks in \cref{tab:2d-point-tracking}. Our method achieves the best average $\delta_{\text{avg}}$ and performs on par with the dedicated 2D tracker Track-On2~\cite{Aydemir2025TrackOn2EO} on AJ and OA.

\begin{figure}[t]
    \centering
    \includegraphics[width=\linewidth]{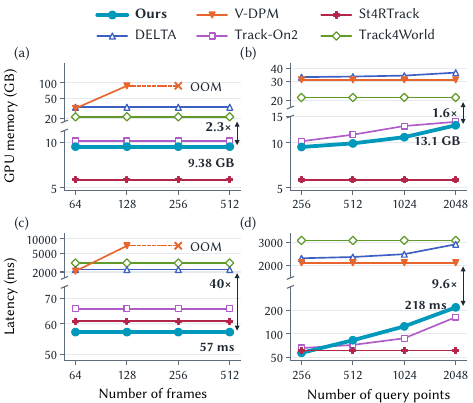}
    \caption{\small \textbf{Complexity Analysis.} GPU memory (top) and latency (bottom) versus the number of frames (left, 256 points) and query points (right, 64 frames).}
    \label{fig:complexity}
    \vspace{-0.1in}
\end{figure}

\subsection{Complexity Analysis}
\label{sec:complexity}

\Cref{fig:complexity} compares GPU memory and latency (time to the first output) on the same NVIDIA Hopper GPU with RGB-only inputs at $384 \times 512$. With 256 query points, our memory (9.38 GB) and latency (57 ms) stay constant from 64 to 512 frames, sustaining 17 FPS in steady-state streaming, whereas full-sequence methods quickly run out of memory and sliding-window methods are far more costly. Among causal trackers, StreamTAP3D runs at a cost comparable to Track-On2 (with Abot-Recon) and requires more memory only than the memory-free St4RTrack, while being substantially more accurate than both (\cref{tab:worldcoord-tracking}).

\subsection{Ablations}
\label{sec:ablations}

\begin{table}[tb!]
\centering
\captionsetup{skip=2pt}
\caption{\textbf{Ablation Study on the Memory Mechanism.} We report $\text{APD}_{\text{3D}}\uparrow$ across the four WorldTrack subsets.}
\label{tab:ablation-memory}
\setlength{\tabcolsep}{6pt}
\renewcommand{\arraystretch}{1.0}
\resizebox{\columnwidth}{!}{%
    \begin{tabular}{@{}lccccc@{}}
    \toprule
    \textbf{Memory Mechanism} & \textbf{ADT} & \textbf{DR} & \textbf{PO} & \textbf{PStudio} & \textbf{Avg.} \\
    \midrule
    No Memory & 48.15 & 52.67 & 50.92 & 49.26 & 50.25 \\
    Short-Term FIFO Only & 55.24 & 58.49 & 56.62 & 58.73 & 57.27 \\
    Long-Term TTT Only & 53.38 & 57.11 & 53.27 & 55.86 & 54.91 \\
    \textbf{FIFO + TTT (Ours)} & \textbf{56.44} & \textbf{59.38} & \textbf{57.63} & \textbf{59.35} & \textbf{58.20} \\
    \bottomrule
    \end{tabular}%
}
\vspace{-6pt}
\end{table}

\noindent\textbf{Memory Mechanism.}
\Cref{tab:ablation-memory} ablates the short-term FIFO memory and long-term TTT fast-weight memory across the WorldTrack subsets. Introducing either the short-term FIFO cache or the long-term TTT memory individually yields clear improvements over the memory-free baseline across all subsets, while combining both achieves the best overall $\text{APD}_{\text{3D}}$, demonstrating their complementary benefits.

\begin{table}[tb!]
\centering
\captionsetup{skip=2pt}
\caption{\textbf{Ablation Study on the Coupled 2D-3D Architecture.} We report $\text{APD}_{\text{3D}}\uparrow$ across the four WorldTrack subsets.}
\label{tab:ablation-2d3d}
\setlength{\tabcolsep}{6pt}
\renewcommand{\arraystretch}{1.0}
\resizebox{\columnwidth}{!}{%
    \begin{tabular}{@{}lccccc@{}}
    \toprule
    \textbf{Variant} & \textbf{ADT} & \textbf{DR} & \textbf{PO} & \textbf{PStudio} & \textbf{Avg.} \\
    \midrule
    w/o Ray Snapping & 55.11 & 58.01 & 56.15 & 57.22 & 56.62 \\
    w/o Shared Features & 53.64 & 57.18 & 55.14 & 57.93 & 55.97 \\
    3D Training Data Only & 54.37 & 57.24 & 57.16 & 58.81 & 56.90 \\
    Image-Patch Neighborhood & 54.89 & 57.91 & 56.51 & 57.18 & 56.62 \\
    \textbf{Full Model (Ours)} & \textbf{56.44} & \textbf{59.38} & \textbf{57.63} & \textbf{59.35} & \textbf{58.20} \\
    \bottomrule
    \end{tabular}%
}
\vspace{-6pt}
\end{table}

\noindent\textbf{Coupled 2D-3D Architecture.}
We evaluate our coupled 2D-3D design in \cref{tab:ablation-2d3d} by ablating 2D-guided ray snapping, shared visual encoding, and joint 2D-3D training data, and by replacing the world-space nearest-neighbor retrieval with an image-patch retrieval around $\hat{\mathbf{u}}_t^i$. Each variant consistently degrades $\text{APD}_{\text{3D}}$ across all subsets. For the latter, image patches mix foreground and background points across depth discontinuities, whereas world-space retrieval naturally separates them.

\begin{table}[tb!]
\centering
\captionsetup{skip=2pt}
\caption{\textbf{Different Spatial Representations.} We vary the coordinate frame for training and inference, and the predicted update of the 3D branch (3D displacement $\Delta \mathbf{X}$ or depth residual $\Delta d$), all with ray snapping. \emph{Random} applies a randomly sampled sequence of rigid transformations to each training clip. We report $\text{APD}_{\text{3D}}\uparrow$.}
\label{tab:ablation-spatial}
\fontsize{8.58pt}{10.29pt}\selectfont %
\setlength{\tabcolsep}{3pt}
\renewcommand{\arraystretch}{1.0}
\begin{tabular*}{\columnwidth}{@{\extracolsep{\fill}}lllccccc@{}}
    \toprule
    \multicolumn{2}{@{}c}{\textbf{Coord.}} & & & & & & \\
    \cmidrule(r){1-2}
    \textbf{Train} & \textbf{Infer} & \textbf{Update} & \textbf{ADT} & \textbf{DR} & \textbf{PO} & \textbf{PStudio} & \textbf{Avg.} \\
    \midrule
    Camera & Camera & $\Delta \mathbf{X}$ & 52.64 & 56.91 & 54.38 & 56.52 & 55.11 \\
    World & World & $\Delta \mathbf{X}$ & 53.91 & 57.81 & 55.76 & 57.72 & 56.30 \\
    Random & World & $\Delta d$ & 38.74 & 44.16 & 40.62 & 46.96 & 42.62 \\
    Random & Camera & $\Delta \mathbf{X}$ & 56.40 & \textbf{59.54} & 57.02 & \textbf{59.48} & 58.11 \\
    \textbf{Random} & \textbf{World} & $\boldsymbol{\Delta} \mathbf{X}$ & \textbf{56.44} & 59.38 & \textbf{57.63} & 59.35 & \textbf{58.20} \\
    \bottomrule
\end{tabular*}
\vspace{-6pt}
\end{table}

\noindent\textbf{Different Spatial Representations.}
In \cref{tab:ablation-spatial}, training in a fixed camera or world frame underperforms random-frame training, whose model performs similarly when inferred in either frame, suggesting reduced sensitivity to the choice of coordinate frame. In this setting, $\Delta d$ performs worst, because the camera-axis direction is not explicitly provided to the update predictor.

\begin{table}[tb!]
\centering
\captionsetup{skip=2pt}
\caption{\textbf{Ablation Study on the Short-Term Memory Length.} We vary the FIFO cache capacity $B$ and report $\text{APD}_{\text{3D}}\uparrow$ across the four WorldTrack subsets.}
\label{tab:ablation-memory-length}
\fontsize{8.58pt}{10.29pt}\selectfont %
\setlength{\tabcolsep}{4pt}
\renewcommand{\arraystretch}{1.0}
\begin{tabular*}{\columnwidth}{@{\extracolsep{\fill}}lccccc@{}}
    \toprule
    \textbf{Memory Length} & \textbf{ADT} & \textbf{DR} & \textbf{PO} & \textbf{PStudio} & \textbf{Avg.} \\
    \midrule
    $B=3$ & 53.95 & 56.23 & 54.51 & 56.48 & 55.29 \\
    $B=7$ & 55.18 & 57.03 & 55.74 & 58.99 & 56.74 \\
    $\mathbf{B=15}$ \textbf{(Ours)} & \textbf{56.44} & 59.38 & 57.63 & 59.35 & 58.20 \\
    $B=31$ & 56.28 & \textbf{59.58} & \textbf{58.14} & \textbf{59.47} & \textbf{58.37} \\
    \bottomrule
\end{tabular*}
\vspace{-6pt}
\end{table}

\noindent\textbf{Short-Term Memory Length.}
We vary the FIFO capacity $B \in \{3, 7, 15, 31\}$ in \cref{tab:ablation-memory-length}. Accuracy improves up to $B=15$, beyond which enlarging the cache brings no clear gain but higher per-frame cost; we thus adopt $B=15$.

\begin{table}[tb!]
\centering
\captionsetup{skip=2pt}
\caption{\textbf{Dependence on Geometry Quality.} We report $\text{APD}_{\text{3D}}\uparrow$ on DR and PO, using the fine-tuned Abot-Recon or ground-truth (GT) depth and camera parameters for all methods.}
\label{tab:geometry}
\fontsize{8.58pt}{10.29pt}\selectfont %
\setlength{\tabcolsep}{4pt}
\renewcommand{\arraystretch}{1.0}
\begin{tabular*}{\columnwidth}{@{\extracolsep{\fill}}lcccccc@{}}
    \toprule
    & \multicolumn{3}{c}{\textbf{Fine-tuned Abot-Recon}} & \multicolumn{3}{c}{\textbf{GT Geometry}} \\
    \cmidrule(lr){2-4} \cmidrule(lr){5-7}
    \textbf{Method} & \textbf{DR} & \textbf{PO} & \textbf{Avg.} & \textbf{DR} & \textbf{PO} & \textbf{Avg.} \\
    \midrule
    SPOT + lift & 57.06 & 52.29 & 54.68 & 58.77 & 59.01 & 58.89 \\
    Track-On2 + lift & 56.39 & 51.84 & 54.12 & 59.62 & 61.84 & 60.73 \\
    \textbf{Ours} & \textbf{59.38} & \textbf{57.63} & \textbf{58.51} & \textbf{87.69} & \textbf{70.56} & \textbf{79.13} \\
    \bottomrule
\end{tabular*}
\vspace{-6pt}
\end{table}

\noindent\textbf{Dependence on Geometry Quality.}
With ground-truth geometry on DR and PO (\cref{tab:geometry}), the lifted 2D trackers show limited improvement, whereas StreamTAP3D improves substantially, consistent with the observation in TAPIP3D~\cite{Zhang2025TAPIP3DTA}, suggesting that our accuracy is bounded by reconstruction and will benefit from future advances.

\begin{table}[tb!]
\centering
\captionsetup{skip=2pt}
\caption{\textbf{Different Reconstruction Backbones.} We report $\text{APD}_{\text{3D}}\uparrow$ of our tracker paired with different streaming reconstructors on the four WorldTrack subsets. ``Frames'' denotes evaluation over the first 64 frames or on full-length sequences.}
\label{tab:ablation-backbone}
\fontsize{8.58pt}{10.29pt}\selectfont %
\setlength{\tabcolsep}{3.5pt}
\renewcommand{\arraystretch}{1.0}
\begin{tabular*}{\columnwidth}{@{\extracolsep{\fill}}llccccc@{}}
    \toprule
    \textbf{Frames} & \textbf{Backbone} & \textbf{ADT} & \textbf{DR} & \textbf{PO} & \textbf{PStudio} & \textbf{Avg.} \\
    \midrule
    \multirow{3}{*}{64} & Abot-Recon & 56.44 & \textbf{59.38} & \textbf{57.63} & \textbf{59.35} & 58.20 \\
    & LingBot-Map & 66.76 & 48.69 & 46.07 & 53.69 & 53.80 \\
    & IGGT4D & \textbf{78.58} & 55.21 & 55.75 & 59.06 & \textbf{62.15} \\
    \midrule
    \multirow{3}{*}{Full} & Abot-Recon & 42.29 & \textbf{58.28} & 54.47 & 55.89 & 52.73 \\
    & LingBot-Map & 56.42 & 49.28 & 45.64 & 52.63 & 50.99 \\
    & IGGT4D & \textbf{64.98} & 55.52 & \textbf{55.49} & \textbf{57.05} & \textbf{58.26} \\
    \bottomrule
\end{tabular*}
\vspace{-6pt}
\end{table}

\noindent\textbf{Different Reconstruction Backbone.}
\Cref{tab:ablation-backbone} pairs our tracker with different streaming reconstructors. IGGT4D~\cite{zou2026iggt4d} achieves the highest overall accuracy, mainly on ADT, but its cost grows with video length. Among constant-cost reconstructors, Abot-Recon~\cite{han2026abotrecon} outperforms LingBot-Map~\cite{Chen2026LingBotMapGC} on average and is thus adopted by default (cost analysis in the supplementary material).

\begin{table}[tb!]
\centering
\captionsetup{skip=2pt}
\caption{\textbf{Fast-Weight Memory on Long Sequences.} We report $\text{APD}_{\text{3D}}\uparrow$ and survival rate (Surv.$\uparrow$) on the official DR and PO test sets (900 frames). Survival rate measures the fraction of the video a track covers before it is first lost (see supp.).}
\label{tab:ablation-ttt-long}
\fontsize{8.58pt}{10.29pt}\selectfont %
\setlength{\tabcolsep}{4pt}
\renewcommand{\arraystretch}{1.0}
\begin{tabular*}{\columnwidth}{@{\extracolsep{\fill}}lcccccc@{}}
    \toprule
    & \multicolumn{2}{c}{\textbf{DR}} & \multicolumn{2}{c}{\textbf{PO}} & \multicolumn{2}{c}{\textbf{Avg.}} \\
    \cmidrule(lr){2-3} \cmidrule(lr){4-5} \cmidrule(lr){6-7}
    \textbf{Memory} & $\text{APD}_{\text{3D}}$ & Surv. & $\text{APD}_{\text{3D}}$ & Surv. & $\text{APD}_{\text{3D}}$ & Surv. \\
    \midrule
    FIFO Only & 47.67 & 66.13 & 32.74 & 40.64 & 40.21 & 53.39 \\
    \textbf{Ours} & \textbf{48.91} & \textbf{67.62} & \textbf{34.88} & \textbf{43.26} & \textbf{41.90} & \textbf{55.44} \\
    \bottomrule
\end{tabular*}
\vspace{-6pt}
\end{table}

\noindent\textbf{Fast-Weight Memory on Long Sequences.}
To further evaluate the fast-weight memory on longer sequences, we ablate it on the first 900 frames of the official DR and PO test sets in \cref{tab:ablation-ttt-long}. The fast-weight memory improves both $\text{APD}_{\text{3D}}$ and survival rate, supporting the benefit of fast-weight memory on long sequences.

\section{Conclusion}
\label{sec:conclusion}
We presented StreamTAP3D, a streaming 3D point tracker that processes monocular RGB video frame by frame with zero lookahead, using a hybrid long-short-term memory for temporal context under a bounded budget and a coupled 2D-3D architecture to stabilize 3D refinement under noisy online geometry. It achieves competitive accuracy at practical efficiency, enabling live 3D point tracking.

\noindent\textbf{Limitations.}
Our framework has two main limitations. First, its accuracy depends on the quality of streaming reconstruction: observing only past frames, streaming reconstructors are more prone to drift than offline models such as VGGT~\cite{Wang2025VGGTVG}, which propagates into the tracked trajectories. We expect our method to benefit from future advances in streaming reconstruction. Second, the 2D-3D coupling is one-directional: 2D predictions refine 3D tracks, but 3D cannot in turn correct the 2D branch, so 2D failures propagate to 3D. A promising direction is bidirectional 2D-3D refinement, in which each branch corrects the other.

{
    \small
    \bibliographystyle{ieeenat_fullname}
    \bibliography{main}
}

\clearpage
\maketitlesupplementary
\setcounter{section}{0}
\renewcommand{\thesection}{\Alph{section}}
\renewcommand{\theHsection}{supp.\Alph{section}}
\setcounter{figure}{0}
\renewcommand{\thefigure}{S\arabic{figure}}
\setcounter{table}{0}
\renewcommand{\thetable}{S\arabic{table}}
\setcounter{equation}{0}
\renewcommand{\theequation}{S\arabic{equation}}

This supplementary material provides additional implementation details of StreamTAP3D (\cref{sec:supp-impl}), details of the Online Geometry Predictor (\cref{sec:supp-geometry}), and details of the training data (\cref{sec:supp-data}), as well as additional experiments, including an ablation on our synthetic training data and a cost analysis of different reconstruction backbones (\cref{sec:supp-ablation}).

\section{Implementation Details}
\label{sec:supp-impl}

\subsection{Network Architecture}
\label{sec:supp-arch}

\begin{table}[tb!]
\centering
\captionsetup{skip=2pt}
\caption{\textbf{Hyperparameters of StreamTAP3D.} Values shared by both branches are listed once.}
\label{tab:supp-hyper}
\footnotesize
\setlength{\tabcolsep}{3.5pt}
\renewcommand{\arraystretch}{1.0}
\begin{tabular*}{\columnwidth}{@{\extracolsep{\fill}}llcc@{}}
    \toprule
    \textbf{Module} & \textbf{Hyperparameter} & \textbf{2D} & \textbf{3D} \\
    \midrule
    \multirow{3}{*}{Visual Encoder} & Feature dim. & \multicolumn{2}{c}{128} \\
    & Base stride $s$ & \multicolumn{2}{c}{4} \\
    & Pyramid levels $S$ & \multicolumn{2}{c}{4} \\
    \midrule
    \multirow{3}{*}{Correlation} & Support & $7 \times 7$ patch & $J = 32$ neighbors \\
    & Context & $7 \times 7$ patch & $J = 32$ neighbors \\
    & Dim. per level & 256 & 128 \\
    \midrule
    \multirow{6}{*}{Refiner} & Input dim. & 1110 & 777 \\
    & Layers $L$ / width $C$ & \multicolumn{2}{c}{3 / 384} \\
    & Heads / MLP ratio & \multicolumn{2}{c}{8 / 4} \\
    & Virtual tracks & \multicolumn{2}{c}{64} \\
    & FIFO capacity $B$ & \multicolumn{2}{c}{15} \\
    & Iterations $M$ & \multicolumn{2}{c}{4} \\
    \midrule
    \multirow{3}{*}{TTT Memory} & Heads $\times$ dim. $d$ & \multicolumn{2}{c}{$12 \times 32$} \\
    & Initial step size $\beta$ & \multicolumn{2}{c}{0.1} \\
    & Write input dim. & 1622 & 1223 \\
    \bottomrule
\end{tabular*}
\vspace{-6pt}
\end{table}

\Cref{tab:supp-hyper} lists the hyperparameters of both branches. Below, $\gamma(\cdot)$ denotes a NeRF-style sinusoidal encoding.

\noindent\textbf{Visual Encoder.}
We adopt the CNN encoder of CoTracker3~\cite{Karaev2025CoTracker3SA}, which outputs $128$-dimensional features at stride $4$. The $S = 4$ pyramid levels are obtained by average pooling. The encoder runs once per frame and its features are read by both branches.

\noindent\textbf{Correlation.}
The 2D correlation follows CoTracker3: at each level, the inner products between a $7 \times 7$ query patch and a $7 \times 7$ patch around the current estimate are embedded by an MLP. For the 3D correlation, the point map is downsampled to each pyramid level and normalized by the statistics of the first-frame point cloud. The support and context tokens each contain $J = 32$ nearest neighbors, and each token adds an embedding of its relative 3D offset to its feature. A bidirectional N2N cross-attention block followed by attention pooling yields one embedding per level.

\noindent\textbf{Dual-Memory Refiner.}
Both branches use a transformer with $L = 3$ layers of width $C = 384$. Space Attention exchanges information through $64$ virtual tracks. In \cref{eq:token}, $\mathbf{h}_\tau^i$ contains the correlation embedding and visibility logit (plus a confidence logit in 2D), and the 3D token additionally encodes the projected pixel location. When a track has fewer than $B$ past frames, the empty FIFO slots repeat its earliest entry.

\subsection{Long-Term Fast-Weight Memory}
\label{sec:supp-ttt}

\noindent\textbf{Write Token.}
After the final iteration of frame $t$, we resample features at the converged prediction $\mathbf{p}_t^i$. The write token $\hat{\mathbf{z}}_t^i$ is obtained by an MLP from the encoded position $\gamma(\mathbf{p}_t^i)$ and motion $\gamma(\mathbf{p}_t^i - \mathbf{p}_{t-1}^i)$, the visibility logit, the recomputed correlation embedding, and the local appearance feature. The memory is written at every frame, including occluded ones; since the predicted visibility is part of the write input, the learned key, target and step size can adapt to occlusion.

\noindent\textbf{Parameterization.}
The fast weights of each layer are split into $12$ heads of size $d \times d$ with $d = 32$, initialized from a learnable state $\mathbf{W}_0$ shared by all tracks. Keys are $\ell_2$-normalized and the step size is predicted as $\beta_t^i = \sigma(\cdot) \in (0, 1)$ (initialized to $0.1$), so each update in \cref{eq:ttt-write} moves $\mathbf{W}\mathbf{k}_t^i$ only part of the way towards $\mathbf{y}_t^i$. The read-out is added to $\mathbf{z}_\tau^i$ through a gate initialized to zero, so the memory initially leaves the refiner unchanged.

\subsection{Training Details}
\label{sec:supp-training}

\noindent\textbf{Data Augmentation.}
Each training clip is augmented on the image plane while keeping the 3D scene fixed. We randomly pad each side by $0$--$25$ pixels and rescale the frames by a factor in $[0.9, 1.25]$, with an additional small per-frame scale jitter ($\delta = 0.05$). The clip is then cropped to $384 \times 512$ around the projections of the tracks visible in the first frame, with the crop shifted by up to $15$ pixels per frame. Ground-truth depth is resampled with nearest-neighbor interpolation and the intrinsics are updated accordingly, so these operations change only the depth maps and intrinsics, not the world coordinates of the 3D points. We further apply horizontal flipping and temporal reversal, each with probability $0.5$, which transform the extrinsics, 3D trajectories and visibilities consistently.

\noindent\textbf{Random World Frame.}
To make the 3D branch agnostic to the camera trajectory (\cref{tab:ablation-spatial}), we re-express each training clip in a randomly sampled world frame. We generate a smooth random camera trajectory whose linear and angular velocities are initialized uniformly in $[0, 0.03]$ (relative to the scene scale) and $[0, 0.1]$, and are then perturbed at every frame by random noise of magnitude $0.015$ and $0.05$, respectively. The camera pose of every frame is replaced by this trajectory, and all 3D points of the frame, including the trajectories and the query points, are transformed by the same pose. Since points and camera move together, all projections and depth maps are unchanged; only the world-space trajectories and camera motion differ.

\noindent\textbf{Loss and Optimization.}
In \cref{eq:loss}, we set $\lambda_{\text{traj}} = 0.0025$ and $\lambda_{\text{bce}} = 0.0075$, and the 2D $L_1$ term in $\mathcal{L}_{\text{traj}}$ is weighted equally with the 3D term. We use AdamW with a peak learning rate of $10^{-4}$ and a weight decay of $5 \times 10^{-4}$, under a one-cycle schedule with cosine annealing: the learning rate warms up from $4 \times 10^{-6}$ to $10^{-4}$ over the first $5\%$ of the steps and then decays to $10^{-8}$. The schedule is restarted at each stage of the sequence-length curriculum, spanning its $50$k, $10$k and $10$k iterations, respectively. Training uses BF16 mixed precision and gradient clipping at norm $10$, and a step is skipped on all GPUs if the loss or gradients are non-finite on any of them. We use a batch size of $1$ clip per GPU.

\subsection{Evaluation Details}
\label{sec:supp-eval}

\noindent\textbf{Survival Rate.}
We report the survival rate in \cref{tab:ablation-ttt-long} to evaluate the fast-weight memory on long sequences. Since the official DR and PO test sets used in this table contain a very large number of query points, we select $2048$ queries per sequence by farthest point sampling for evaluation. Following PointOdyssey~\cite{Zheng2023PointOdysseyAL}, it measures how long a track is followed before it is first lost. Given a distance threshold $\tau$, track $i$ fails at frame $t$ if the point is visible with a valid ground truth and its 3D error $d_t^i = \|\hat{\mathbf{X}}_t^i - \mathbf{X}_t^i\|_2$, computed after the same alignment as $\text{APD}_{\text{3D}}$, exceeds $\tau$ or is non-finite. Once a track fails, it is considered lost for all subsequent frames, while occluded or invalid frames do not interrupt survival. This gives a binary survival mask
\begin{equation}
    s_t^i(\tau) = \prod_{t' \leq t} \Big( 1 - \mathbb{1}\big[v_{t'}^i = 1 \,\wedge\, \neg (d_{t'}^i \leq \tau)\big] \Big),
\end{equation}
where $v_{t'}^i = 1$ indicates a visible frame with valid ground truth and $\neg(d \leq \tau)$ also covers non-finite errors. The score of a sequence is the mean of $s_t^i(\tau)$ over all frames and tracks, multiplied by $100$ and averaged over $\tau \in \{0.1, 0.3, 0.5, 1.0, 1.5, 2.5\}$. We report the average over all sequences.

\noindent\textbf{WorldTrack Protocol.}
We follow the evaluation protocol of Track4World~\cite{Lu2026Track4WorldFW} on WorldTrack. We track the points visible in the first frame of each sequence. The queries are lifted to world coordinates and kept there, and the 3D branch directly predicts world-space trajectories. For scoring, the ground-truth and predicted trajectories are transformed into the first-frame camera coordinates of their respective geometry (ground-truth or predicted). The predictions are then aligned to the ground truth by an isotropic scale and an $xyz$ translation estimated under an $L_1$ objective, followed by the median scale alignment of TAP-Vid3D~\cite{Koppula2024TAPVid3DAB}. $\text{APD}_{\text{3D}}$ and $\text{AJ}_{\text{3D}}$ are computed with fixed distance thresholds of $0.01$, $0.04$, $0.16$, $0.64$ and $2.56$\,m, and a point is predicted as visible if its visibility probability exceeds $0.5$.

\section{Online Geometry Predictor}
\label{sec:supp-geometry}

\noindent\textbf{Training Data and Schedule.}
We initialize ABot-Recon~\cite{han2026abotrecon} from its released weights and fine-tune it for 22,000 optimizer updates on 16 GPUs.
The initial mixture contains prepared subsets of 13 sources: BlinkVision Indoor and Outdoor, Virtual KITTI~2, Dynamic Replica, TartanAir, MVS-Synth, ScanNet, Spring, ScanNet++, PointOdyssey, Kubric, GTA-SfM, and OmniWorld.
We sample clips at $280\times504$ or $378\times518$ pixels (height by width) and adapt the per-GPU batch size to the source, resolution, and clip length.
Clips contain 16, 24, 32, 48, or 64 frames, with Kubric restricted to 16 or 24.
The tracker's input resolution is $384\times512$.
The reconstructor's local causal window contains the current frame and up to 11 preceding frames.
We freeze the image encoder, disable the confidence branch, and optimize all remaining parameters using AdamW with a one-cycle schedule, a peak learning rate of $10^{-6}$, and weight decay of $0.05$.
We maintain an exponential moving average (EMA) with decay $0.999$ and validate every 500 updates.

\noindent\textbf{Geometric Supervision.}
We back-project ground-truth depth with the camera intrinsics to obtain a target 3D point at each valid pixel.
After aligning one scale per clip, we minimize the inverse-depth-weighted $\ell_1$ distance between predicted and target camera-space points.
We also supervise surface normals on TartanAir, PointOdyssey, ScanNet, and Virtual KITTI~2.
Intrinsics enter through back-projection and receive no separate regression loss.
The camera loss supervises relative poses for frame pairs up to 11 time steps apart.
It combines a Huber translation loss (weight 100), a rotation-angle loss (weight 1), and penalties on the magnitude and temporal variation of the rotation correction (weight $10^{-3}$ each).
We add the camera loss to the point and normal losses with weight $0.1$.

\section{Training Data}
\label{sec:supp-data}

\begin{table}[tb!]
\centering
\captionsetup{skip=2pt}
\caption{\textbf{Training Data Mixture.} Domain, annotation type and sampling ratio of each dataset. For 2D-only datasets, only the 2D branch is supervised.}
\label{tab:supp-mixture}
\footnotesize
\setlength{\tabcolsep}{3.5pt}
\renewcommand{\arraystretch}{1.0}
\begin{tabular*}{\columnwidth}{@{\extracolsep{\fill}}lccc@{}}
    \toprule
    \textbf{Dataset} & \textbf{Domain} & \textbf{Annotation} & \textbf{Ratio} \\
    \midrule
    Kubric~\cite{Greff2022KubricAS} & Synthetic & 3D & 0.15 \\
    DynamicReplica~\cite{Karaev2023DynamicStereoCD} & Synthetic & 3D & 0.15 \\
    PointOdyssey~\cite{Zheng2023PointOdysseyAL} & Synthetic & 3D & 0.15 \\
    Stereo4D~\cite{Jin2024Stereo4DLH} & Real & 2D & 0.20 \\
    FlyingThings3D~\cite{Mayer2015ALD} & Synthetic & 2D & 0.05 \\
    Driving~\cite{Mayer2015ALD} & Synthetic & 2D & 0.05 \\
    Monkaa~\cite{Mayer2015ALD} & Synthetic & 2D & 0.05 \\
    Spring~\cite{Mehl2023SpringAH} & Synthetic & 2D & 0.05 \\
    Ours & Synthetic & 3D & 0.15 \\
    \bottomrule
\end{tabular*}
\vspace{-6pt}
\end{table}

\Cref{tab:supp-mixture} lists the sampling ratio of each training dataset. For every clip, the frame stride is sampled from $\{1, 2, 3\}$ among the values for which the source sequence is long enough to provide the required number of frames; \eg, a $64$-frame clip from a $90$-frame sequence always uses stride $1$.
We will release the rendering scripts of our synthetic dataset.

\section{Additional Experiments}
\label{sec:supp-ablation}

\begin{table}[tb!]
\centering
\captionsetup{skip=2pt}
\caption{\textbf{Ablation on Our Synthetic Dataset.} We report $\text{APD}_{\text{3D}}\uparrow$ on the four WorldTrack subsets when training with or without our synthetic 3D tracking dataset.}
\label{tab:supp-ablation-data}
\footnotesize
\setlength{\tabcolsep}{3.5pt}
\renewcommand{\arraystretch}{1.0}
\begin{tabular*}{\columnwidth}{@{\extracolsep{\fill}}lccccc@{}}
    \toprule
    \textbf{Training Data} & \textbf{ADT} & \textbf{DR} & \textbf{PO} & \textbf{PStudio} & \textbf{Avg.} \\
    \midrule
    w/o Synthetic Dataset & \textbf{56.61} & 59.02 & 56.93 & 57.85 & 57.60 \\
    \textbf{Full (Ours)} & 56.44 & \textbf{59.38} & \textbf{57.63} & \textbf{59.35} & \textbf{58.20} \\
    \bottomrule
\end{tabular*}
\vspace{-6pt}
\end{table}

\noindent\textbf{Synthetic Training Data.}
\Cref{tab:supp-ablation-data} evaluates the contribution of our synthetic 3D tracking dataset by retraining the tracker without it, keeping all other data and training settings unchanged. Removing the synthetic data lowers the average $\text{APD}_{\text{3D}}$ from 58.20 to 57.60, mainly due to a drop of 1.50 on PStudio, while the other subsets change only slightly. This indicates that our synthetic data provides a moderate overall gain, concentrated on PStudio.

\begin{figure}[t]
    \centering
    \includegraphics[width=\linewidth]{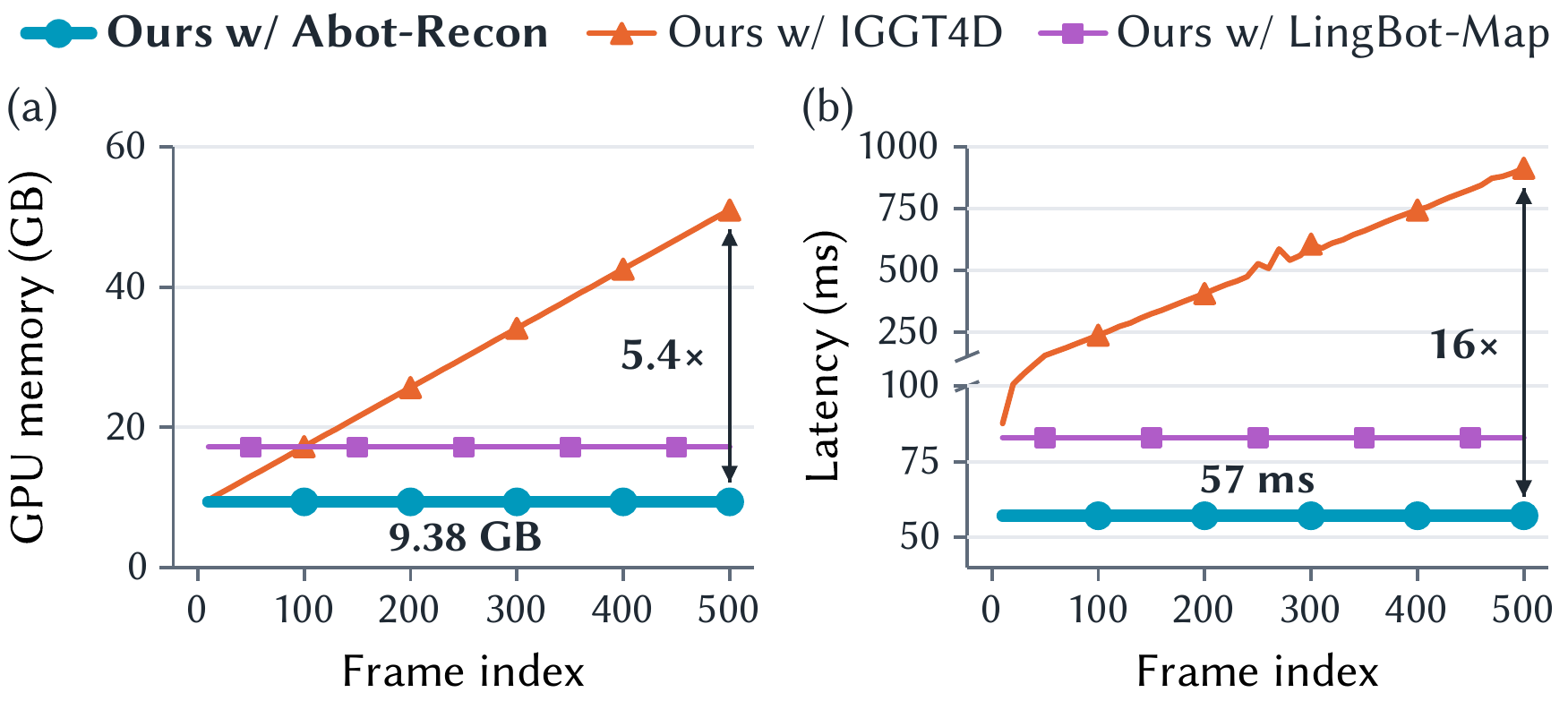}
    \caption{\small \textbf{Cost of Different Online Geometry Predictors.} GPU memory (a) and per-frame latency (b) of the full system when tracking 256 points over a 500-frame $384 \times 512$ video.}
    \label{fig:supp-recon-cost}
\end{figure}

\noindent\textbf{Cost of Different Reconstruction Backbones.}
\label{sec:supp-recon-cost}
We compare the cost of the full system when paired with different streaming reconstructors in \cref{fig:supp-recon-cost}, tracking 256 points on a 500-frame $384 \times 512$ video. IGGT4D~\cite{zou2026iggt4d} achieves the highest overall accuracy (\cref{tab:ablation-backbone}), but its GPU memory and latency grow linearly over time, from 9.64\,GB and 88\,ms at frame 10 to 51.04\,GB and 915\,ms (1.1 FPS) at frame 500. In contrast, both Abot-Recon~\cite{han2026abotrecon} and LingBot-Map~\cite{Chen2026LingBotMapGC} keep memory and latency constant, at 9.38\,GB and 57\,ms for Abot-Recon and 17.2\,GB and 83\,ms for LingBot-Map. Since Abot-Recon is also more accurate than LingBot-Map on average, we adopt it by default, accepting a modest loss in accuracy relative to IGGT4D for the bounded cost and real-time speed required by long-running streams.

\end{document}